\documentclass[conference]{IEEEtran}

\usepackage{placeins}

\usepackage{amsmath,amssymb}
\usepackage{booktabs}
\usepackage{array}
\usepackage{multirow}
\usepackage{graphicx}
\usepackage{url}
\usepackage{enumitem}

\usepackage{algorithm}
\usepackage{algpseudocode}

\floatstyle{ruled}
\restylefloat{algorithm}

\algrenewcommand{\algorithmicrequire}{\textbf{Input:}}
\algrenewcommand{\algorithmicensure}{\textbf{Output:}}
\algrenewcommand{\alglinenumber}[1]{\normalfont #1:}
\algrenewcommand{\algorithmicindent}{1.2em}

\usepackage[hidelinks]{hyperref}

\IEEEoverridecommandlockouts

\title{AGGR-Mamba: Mamba-based Network with Attention-Guided Gated Residual Refinement for Undersampled MRI Reconstruction}

\author{
\IEEEauthorblockN{
Yue Zhuo\textsuperscript{1},
Wenbin Deng\textsuperscript{1},
Lingjun Liu\textsuperscript{2},
Xuhang Chen\textsuperscript{2},
Zhonghua Xie\textsuperscript{2,*}
}
\IEEEauthorblockA{
\begin{tabular}{c}
\textsuperscript{1}School of Mathematics and Statistics, Huizhou University, Huizhou, China\\
\textsuperscript{2}School of Computer Science and Engineering, Huizhou University, Huizhou, China
\end{tabular}
}
\thanks{*Corresponding author: eexie@qq.com. This work was supported by
the National Natural Science Foundation of China (Grant No. 62001184),
the Guangdong Basic and Applied Basic Research Foundation (Grant No.
2025A1515110445), the Projects of Educational Commission of Guangdong
Province (Grant Nos. 2025ZDZX1038, 2025KTSCX49, and 2023ZDZX1025),
the Huizhou University Independent Innovation Capability Enhancement
Plan Project (Grant No. HZU202535), and the College Students' Innovative
Entrepreneurial Training Plan Program (Grant Nos. 202410577008 and
202610577034).}
}

\begin{document}
\maketitle

\begin{abstract}
Undersampled magnetic resonance imaging (MRI) reconstruction aims to recover high-quality images from partially acquired k-space data, thereby reducing scan time. However, missing frequency-domain measurements often lead to residual aliasing artifacts, blurred edges, and loss of local structural details. In this paper, we propose AGGR-Mamba, a lightweight output-stage residual refinement method built upon a Mamba backbone for undersampled brain MRI reconstruction. The Mamba backbone first produces an initial reconstruction, and the proposed refinement modules then correct the remaining local errors and high-frequency artifacts in a controlled residual manner.Specifically, an Attention Refine Head is introduced to recalibrate local anatomical structures, edge textures, and residual-error regions through channel and spatial attention. In addition, a GDFN-style gated residual refinement branch is designed to selectively preserve informative residual features while suppressing redundant or unstable responses. By combining attention-based recalibration with gated residual selection, AGGR-Mamba improves local structure recovery without substantially disturbing the backbone reconstruction. Experiments are conducted on the IXI brain MRI dataset with T1- and T2-weighted images under acceleration factors of $R=4$ and $R=8$. The results show that AGGR-Mamba achieves competitive PSNR and SSIM performance across different reconstruction settings, demonstrating the effectiveness of the proposed output-stage refinement strategy.
\end{abstract}

\begin{IEEEkeywords}
Magnetic resonance image reconstruction; undersampling; Mamba; IXI dataset
\end{IEEEkeywords}

\begin{figure}[!t]
\centering
\includegraphics[width=\columnwidth]{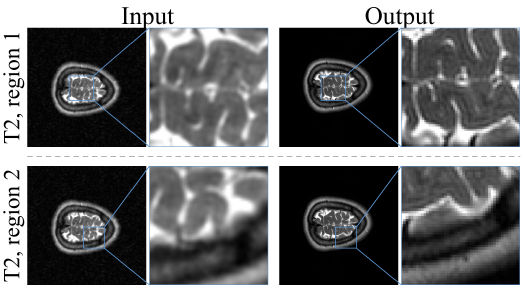}
\par\vspace{-6pt}
\setlength{\abovecaptionskip}{3pt}
\caption{Illustration of the undersampled MRI reconstruction task: undersampled input (left) and AGGR-Mamba reconstruction (right) on IXI-T2 with $R=4$. The two rows highlight different regions of the same slice; blue boxes and lines indicate the corresponding enlarged regions.}
\label{fig:input_output_regions}
\end{figure}

\section{INTRODUCTION}

Magnetic resonance imaging (MRI) is an important medical imaging technique that provides high soft-tissue contrast, rich structural information, and no exposure to ionizing radiation. Owing to these advantages, it has been widely used in the analysis of brain disorders, clinical diagnosis, and medical research~\cite{1}. However, MRI data acquisition is typically time-consuming. Long acquisition times can compromise clinical workflow efficiency, increase patient discomfort, and introduce motion-related artifacts caused by patient movement, which may adversely affect subsequent diagnosis and image analysis~\cite{2}. To improve MRI acquisition efficiency, a commonly used strategy is to undersample the data in k-space. By acquiring fewer k-space measurements, undersampling can reduce scan time, but the resulting lack of complete frequency-domain information makes reconstructed images susceptible to aliasing artifacts, loss of textural details, and blurred edges~\cite{3}.

Figure~\ref{fig:input_output_regions} illustrates the undersampled MRI reconstruction task using an undersampled input and the corresponding AGGR-Mamba output, with two local regions enlarged for comparison.

Although conventional compressed sensing MRI methods are supported by a solid theoretical foundation~\cite{4,5}, they typically rely on manually designed regularization terms and require the tuning of multiple hyperparameters~\cite{6}, while also suffering from relatively low computational efficiency~\cite{7}. Driven by advances in deep learning~\cite{8,9}, CNN-based approaches have been widely explored for learning the reconstruction mapping between undersampled and fully sampled MR images~\cite{10}. CNN-based methods are well suited for local feature extraction, enabling effective recovery of edges and textural details. However, their reliance on local convolutional operations limits their capacity to capture long-range dependencies and global anatomical structures. GAN-based approaches were later explored to enhance the perceptual quality of reconstructed images~\cite{11,12}, but the adversarial learning process may introduce training instability and produce anatomically implausible structures.

To better capture global contextual dependencies, Transformer-based approaches have been explored for MRI reconstruction~\cite{13}. These methods exploit self-attention to capture long-range dependencies, thereby partially compensating for the inherent limitation of CNNs in global context modeling. However, the self-attention mechanism in Transformers incurs high computational complexity, often resulting in substantial memory footprint and computational burden in high-resolution medical image reconstruction~\cite{14}. In recent years, state space models (SSMs), exemplified by Mamba~\cite{15}, have introduced a new paradigm for long-sequence modeling. Compared with conventional Transformers, Mamba is capable of modeling long-range dependencies at significantly lower complexity and has been progressively integrated into medical image reconstruction tasks.

Despite the strengths of Mamba-type reconstruction frameworks in global structural modeling and computational efficiency~\cite{16,17}, the restoration of local details and edge-structure preservation at the output stage in undersampled MRI reconstruction can still be further improved. On the one hand, the reconstructed images may still retain local residual artifacts and high-frequency errors. On the other hand, fine anatomical edges, texture transition regions, and low-contrast structures are susceptible to blurring or insufficient detail during reconstruction. Moreover, a conventional output Refine Head typically applies a relatively uniform correction across the entire image and struggles to distinguish valid structural details from residual errors and spurious high-frequency noise, which may compromise the stability of the refinement process.

To address these limitations, this paper proposes AGGR-Mamba, an output-stage attention-guided and gated residual refinement method for undersampled MRI reconstruction. The proposed method uses a Mamba backbone to obtain an initial reconstruction and attaches a lightweight refinement branch at the output stage. This branch is designed to compensate for local reconstruction errors, edge blurring, and residual high-frequency artifacts that may remain after the backbone reconstruction. The refinement process combines attention-guided local detail recalibration with gated residual feature selection, allowing the network to enhance informative anatomical structures while suppressing unstable residual responses.

The major contributions of this study are as follows:

\begin{itemize}
\setlength{\itemsep}{0.6em}
\setlength{\parsep}{0.2em}
\setlength{\parskip}{0.2em}

\item \textbf{A lightweight output-stage residual refinement design.}
We introduce a lightweight residual refinement branch after the Mamba backbone. Instead of modifying the main reconstruction process, the proposed design refines the backbone output by compensating for local errors and residual artifacts, leading to improved local structural representation in the reconstructed image.

\item \textbf{An attention-guided local detail recalibration mechanism.}
The proposed Attention Refine Head uses channel and spatial attention to adaptively recalibrate channel and spatial responses associated with local structures and residual-error regions. This design reduces uniform correction over the entire image and makes the refinement process more focused on structurally informative areas.

\item \textbf{A gated residual feature selection strategy.}
Since residual features may contain useful high-frequency details as well as noise and unstable responses, we introduce a GDFN-style gated branch to selectively filter candidate residual features. This mechanism enhances reconstruction-relevant details while suppressing ineffective high-frequency noise and residual artifacts.

\end{itemize}

\section{Related Work}

\subsection{CNN-based MRI Reconstruction}

CNN-based MRI reconstruction methods generally learn a nonlinear mapping from undersampled images to their fully sampled counterparts. Representative architectures include U-Net, deep residual networks, and cascaded convolutional networks~\cite{18,19}. These methods are effective at extracting local features and can recover edges, textures, and fine structural details. However, because convolutional operations mainly rely on local receptive fields, CNN-based models may struggle to restore subtle details and delineate structural boundaries clearly, especially in complex brain anatomy and highly accelerated undersampling scenarios.

\subsection{Physics-guided MRI Reconstruction}

Physics-guided MRI reconstruction methods embed the imaging model of MRI into deep reconstruction networks~\cite{20}. These methods commonly rely on a data consistency module to keep the reconstructed result consistent with the sampled k-space data. By combining physical constraints with the representational capacity of deep networks, these methods tend to provide more stable performance in undersampled MRI reconstruction. Models such as E2E-VarNet~\cite{21} fall into this category. The Mamba backbone used in this study follows the same principle of physics-guided reconstruction, where the backbone reconstruction module and data consistency constraint jointly contribute to improved reconstruction quality.

\subsection{Transformer-based MRI Reconstruction}

Transformer-based methods have also been explored for MRI reconstruction, mainly because self-attention can model long-range dependencies beyond the local receptive field of CNNs. For example, ReconFormer~\cite{22} adopts a recurrent Transformer design with multi-scale attention to improve reconstruction quality. Despite these advantages, the self-attention operation is often costly, and its use in high-resolution image reconstruction can result in heavy GPU memory usage and increased computational burden~\cite{23}.

\subsection{Mamba-based MRI Reconstruction}

Mamba-based methods model long-range dependencies through state space models. Compared with Transformer-based models, their linear-complexity sequence modeling is more efficient for long inputs, which is beneficial for high-resolution medical image reconstruction. In MRI reconstruction, Mamba-based designs are commonly combined with data consistency constraints so that the network can exploit both learned feature representations and the acquired k-space measurements~\cite{17}.

Although these methods have shown strong potential in global context modeling and physics-guided reconstruction, most existing designs mainly focus on the main reconstruction stages. The output stage is less explicitly explored, especially for local error compensation, edge-detail preservation, and residual high-frequency artifact suppression. Motivated by this observation, AGGR-Mamba keeps the Mamba backbone as the main reconstruction component and introduces lightweight output-stage refinement modules to improve local structure recovery and refinement stability.

\section{Methodology}

\subsection{Mamba Backbone}

AGGR-Mamba adopts a Mamba-based reconstruction network as the backbone~\cite{17}. The backbone is organized as a six-stage unrolled reconstruction network. Each reconstruction stage consists of one VSSLayer followed by one Data Consistency (DC) module, where each VSSLayer contains two VSSBlocks. The VSSBlocks are used to model long-range dependencies and global contextual information, while the DC modules enforce consistency between the intermediate reconstruction and the acquired k-space measurements. Therefore, the backbone contains 12 VSSBlocks and six stage-wise DC modules in total. After the six stages, an additional final DC module is applied for data consistency enforcement. Based on the backbone output, AGGR-Mamba further applies the Attention Refine Head and the GDFN-style gated residual refinement branch to compensate for remaining local errors at the output stage.

\begin{figure*}[!t]
\centering
\includegraphics[width=0.95\textwidth]{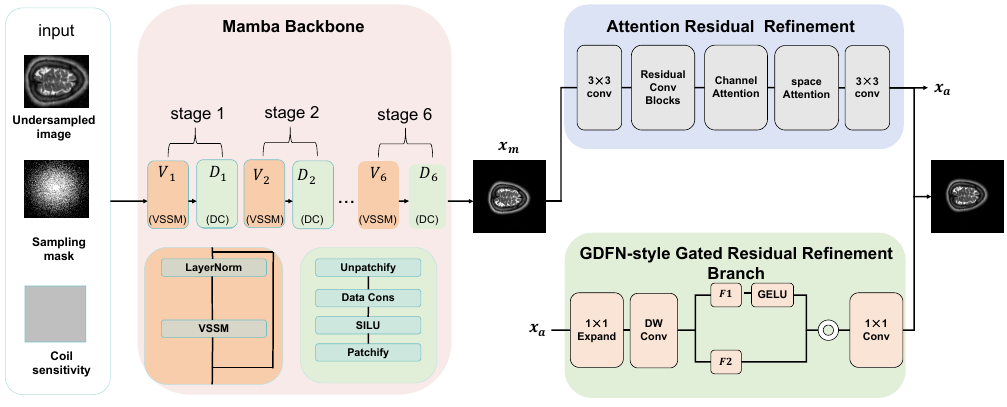}
\vspace{-1em}
\caption{Schematic illustration of the proposed AGGR-Mamba method.}
\label{fig:aggr_mamba_framework}
\vspace{0em}
\end{figure*}

For simplicity, each ``VSSM Block'' illustrated in Fig.~\ref{fig:aggr_mamba_framework} represents a VSSLayer containing two VSSBlocks.

Let $\mathbf{x}$ denote the complex-valued MRI image to be reconstructed, and let $\mathbf{M}$, $\mathbf{C}$, and $\mathbf{n}$ denote the sampling mask, coil sensitivity map, and noise term, respectively. The undersampled MRI acquisition model is written as
\begin{equation}
\mathbf{y}
=
\mathbf{M}\odot\mathcal{F}(\mathbf{C}\mathbf{x})+\mathbf{n},
\end{equation}
where $\mathbf{y}$ denotes the acquired undersampled k-space data, $\mathcal{F}$ denotes the two-dimensional Fourier transform, and $\odot$ denotes element-wise multiplication. The image-domain input reconstructed from the undersampled data is denoted by $\mathbf{x}_{us}$. In the single-coil IXI setting used in this work, the coil sensitivity map is simplified as an identity map, i.e., $\mathbf{C}=\mathbf{1}$.

In the network implementation, each complex MRI image is represented by two channels corresponding to the real and imaginary parts:
\begin{equation}
\mathbf{x}
=
[\mathbf{x}^{(r)},\mathbf{x}^{(i)}]
\in \mathbb{R}^{2\times H\times W}.
\end{equation}
At the $i$-th reconstruction stage, $i=1,\ldots,6$, the input is first processed by the corresponding VSSLayer:
\begin{equation}
\mathbf{z}_{i}
=
\mathcal{V}_{i}(\mathbf{x}_{i-1}),
\end{equation}
and then passed through the DC block:
\begin{equation}
\mathbf{x}_{i}
=
\mathcal{D}_{i}(\mathbf{z}_{i},\mathbf{x}_{us},\mathbf{M},\mathbf{C}).
\end{equation}
Here, $\mathcal{V}_{i}(\cdot)$ denotes the $i$-th VSSLayer and $\mathcal{D}_{i}(\cdot)$ denotes the corresponding DC module. The VSSBlocks within each VSSLayer use multi-directional selective scanning to model spatial interactions among two-dimensional image patches.

For data consistency enforcement, the current prediction is transformed into k-space, and the sampled positions are constrained by the acquired k-space measurements:
\begin{equation}
\mathbf{k}_{pred}
=
\mathcal{F}(\mathbf{C}\mathbf{x}_{pred}),
\quad
\mathbf{k}_{us}
=
\mathcal{F}(\mathbf{C}\mathbf{x}_{us}),
\end{equation}
\begin{equation}
\mathbf{k}_{dc}
=
(1-\mathbf{M})\odot \mathbf{k}_{pred}
+
\mathbf{M}\odot \mathbf{k}_{us},
\end{equation}
\begin{equation}
\mathbf{x}_{dc}
=
\mathbf{C}^{H}\mathcal{F}^{-1}(\mathbf{k}_{dc}).
\end{equation}
Here, $\mathbf{x}_{pred}$ denotes the current complex-valued prediction, $\mathbf{k}_{pred}$ denotes its corresponding k-space representation, $\mathbf{k}_{us}$ denotes the k-space representation associated with the undersampled input, $\mathbf{C}^{H}$ denotes coil combination using the complex-conjugate sensitivity maps, and $\mathbf{x}_{dc}$ denotes the result after data consistency enforcement. In the current IXI experiments, $\mathbf{C}=\mathbf{1}$, so this expression reduces to the single-coil identity-map case implemented in the data loader.

After the unrolled VSSM-DC stages, the Mamba backbone outputs an initial reconstruction:
\begin{equation}
\mathbf{x}_{m}
=
B_{\theta_B}(\mathbf{x}_{us},\mathbf{M},\mathbf{C}),
\end{equation}
where $B_{\theta_B}$ denotes the Mamba backbone and $\theta_B$ denotes its parameters. The output $\mathbf{x}_{m}\in \mathbb{R}^{2\times H\times W}$ contains the real and imaginary components of the reconstructed image. AGGR-Mamba then obtains the final reconstruction by applying the Attention Refine Head and the GDFN-style branch:
\begin{equation}
\hat{\mathbf{x}}
=
G_{\psi}
\left(
A_{\phi}
\left(
B_{\theta_B}(\mathbf{x}_{us},\mathbf{M},\mathbf{C})
\right)
\right),
\end{equation}
where $A_{\phi}$ denotes the Attention Refine Head and $G_{\psi}$ denotes the GDFN-style gated residual refinement branch.

\subsection{Attention Refine Head}

The Mamba backbone can produce a relatively stable initial reconstruction, but residual errors may still remain around local edges, texture-transition regions, and low-contrast structures. To further strengthen local structural representation at the output stage, we attach an Attention Refine Head after the backbone output to perform lightweight feature recalibration on the two-channel reconstruction.

More concretely, the backbone output $\mathbf{x}_{m}$ is first mapped into an intermediate feature space using a $3\times3$ convolution:
\begin{equation}
\mathbf{F}_{0}
=
P_{in}(\mathbf{x}_{m}),
\end{equation}
where $P_{in}$ denotes the input projection layer. Several residual convolutional blocks are then used to extract local contextual features:
\begin{equation}
\mathbf{F}_{r}
=
R_{a}(\mathbf{F}_{0}),
\end{equation}
where $R_a(\cdot)$ denotes the residual convolutional feature extraction operation.

To make the output-stage refinement focus more on structurally informative regions, the Attention Refine Head introduces channel attention and spatial attention at its end. Channel attention generates channel-wise weights using global average pooling and max pooling, while spatial attention produces spatial weights from the average and maximum responses along the channel dimension. This process can be written as
\begin{equation}
\mathbf{F}_{a}
=
\mathrm{SA}
\left(
\mathrm{CA}
\left(
\mathbf{F}_{r}
\right)
\right),
\end{equation}
where $\mathrm{CA}(\cdot)$ and $\mathrm{SA}(\cdot)$ denote channel attention and spatial attention, respectively. Finally, the Attention Refine Head predicts a two-channel residual term, which is added to the backbone output with a learnable scaling factor $s_a$:
\begin{equation}
\mathbf{r}_{a}=P_{out}(\mathbf{F}_{a}),
\end{equation}
\begin{equation}
\mathbf{x}_{a}=\mathbf{x}_{m}+s_a\mathbf{r}_{a}.
\end{equation}
where $P_{out}$ denotes the output projection and $\mathbf{r}_{a}$ is the predicted two-channel residual map. In the implementation, the final output convolution is initialized with zeros, so that the module behaves approximately as an identity mapping at the beginning of optimization. This reduces the initial disturbance introduced by output-stage refinement to the backbone reconstruction.

\subsection{GDFN-style Gated Residual Refinement Branch}

The Attention Refine Head provides an initial recalibration of the output features in both channel and spatial dimensions. On this basis, a GDFN-style gated residual refinement branch is introduced to further screen the candidate residual information. Through gated feature selection, the branch preserves useful detail residuals while filtering redundant or unstable candidate residual responses, thereby reducing their interference with the final reconstruction.

More specifically, the input $\mathbf{x}_{a}$ is first projected by a convolution layer to obtain intermediate features. A $1\times1$ convolution is then used to expand the channel dimension, followed by a depth-wise convolution for local context encoding:
\begin{equation}
\mathbf{F}
=
\mathrm{DWConv}
\left(
P_{e}
\left(
\mathrm{GELU}(P_{g}(\mathbf{x}_{a}))
\right)
\right),
\end{equation}
where $P_g$ denotes the input projection, $P_e$ denotes the feature expansion operation, and $\mathrm{DWConv}$ denotes the depth-wise convolution. The expanded features are then split into two branches along the channel dimension:
\begin{equation}
[\mathbf{F}_{1},\mathbf{F}_{2}]
=
\mathrm{Split}(\mathbf{F}).
\end{equation}
The gated features are obtained by element-wise multiplication:
\begin{equation}
\mathbf{F}_{g}
=
\mathrm{GELU}(\mathbf{F}_{1})
\odot
\mathbf{F}_{2}.
\end{equation}
After that, $\mathbf{F}_{g}$ is mapped back to a two-channel residual through the feature projection and output projection layers:
\begin{equation}
\mathbf{r}_{g}
=
P_{\mathrm{out}}^{g}
\left(P_{\mathrm{proj}}(\mathbf{F}_{g})\right).
\end{equation}
The scaled residual is then added to $\mathbf{x}_{a}$ using the learnable residual coefficient $\alpha$:
\begin{equation}
\hat{\mathbf{x}}
=
\mathbf{x}_{a}
+
\bar{\alpha}\mathbf{r}_{g},
\quad
\bar{\alpha}=\mathrm{clip}(\alpha,0,0.12),
\end{equation}
where $P_{proj}$ denotes the gated-feature projection and $P_{out}^{g}$ denotes the final two-channel output projection. The gating operation $\mathrm{GELU}(\mathbf{F}_{1})\odot\mathbf{F}_{2}$ filters candidate residual responses before they are added to the reconstruction. Its contribution is assessed through the ablation experiments.

\subsection{Training Strategy and Objective Function}

The final AGGR-Mamba optimization adopts a frozen-backbone output-stage fine-tuning strategy. A trained Mamba backbone checkpoint and the existing Attention Refine Head weights are first loaded. During final fine-tuning, the backbone parameters $\theta_B$ are fixed, while the Attention Refine Head parameters $\phi$ and GDFN-style branch parameters $\psi$ remain trainable. The forward path is unchanged: the undersampled input $\mathbf{x}_{us}$, sampling mask $\mathbf{M}$, and coil sensitivity map $\mathbf{C}$ are passed through the backbone, Attention Refine Head, and GDFN-style branch in sequence to produce $\hat{\mathbf{x}}$.

During training, the two real/imaginary channels of the model output $\hat{\mathbf{x}}$ are first converted into a magnitude image:
\begin{equation}
\mathcal{M}(\hat{\mathbf{x}})
=
\sqrt{
\left(\hat{\mathbf{x}}^{(r)}\right)^2
+
\left(\hat{\mathbf{x}}^{(i)}\right)^2
}.
\end{equation}
The training loss is defined as the mean $L_1$ loss computed on the magnitude images:
\begin{equation}
\mathcal{L}
=
\frac{1}{|\mathcal{B}|HW}
\sum_{j=1}^{|\mathcal{B}|}
\left\|
\mathcal{M}(\hat{\mathbf{x}}_{j})
-
\mathbf{x}_{gt,j}
\right\|_{1},
\end{equation}
where $\mathcal{B}$ represents a training batch, with $|\mathcal{B}|$ giving the corresponding batch size. $\mathbf{x}_{gt,j}$ represents the fully sampled target magnitude image corresponding to the $j$-th sample. At the testing stage, the real/imaginary output channels are likewise converted into a magnitude image before calculating PSNR~\cite{24} and SSIM~\cite{25}. In the reported tables, $\mathrm{SSIM}(\%)$ is defined as
\begin{equation}
\mathrm{SSIM}(\%)=100\times \mathrm{SSIM}.
\end{equation}

The trainable parameter set in the final stage is
\begin{equation}
\Omega=\{\phi,\psi\},
\end{equation}
where $\phi$ includes the convolutional, channel-spatial attention, and residual-scaling parameters of the Attention Refine Head, and $\psi$ includes the projection, depth-wise convolution, gating, and residual-scaling parameters of the GDFN-style branch. The frozen backbone satisfies
\begin{equation}
\nabla_{\theta_B}\mathcal{L}=0,
\qquad
\theta_B^{t+1}=\theta_B^{t}.
\end{equation}
Accordingly, the optimization objective is
\begin{equation}
\begin{aligned}
(\phi^{*},\psi^{*})
&=
\arg\min_{\phi,\psi}
\mathcal{L}\left(
G_{\psi}\left(
A_{\phi}\left(
B_{\theta_B}(\mathbf{x}_{us},\mathbf{M},\mathbf{C})
\right)
\right),
\mathbf{x}_{gt}
\right),
\\[-2pt]
&\qquad \theta_B\ \mathrm{fixed}.
\end{aligned}
\end{equation}

The output-stage parameters are optimized using AdamW with a base learning rate of $1\times10^{-3}$ and a minimum learning rate of $1\times10^{-6}$. The batch size is 4, the maximum number of iterations is 5000, and the warmup period is 100 iterations. Validation is performed every 500 iterations, and regular checkpoints are saved every 1000 iterations. For each IXI setting, the best checkpoint is selected according to validation PSNR, while the test set is used only for final performance reporting.

Algorithm~\ref{alg:output_stage_finetuning} summarizes the final-stage optimization and evaluation procedure for each IXI setting.

\begin{algorithm}[t]
\caption{AGGR-Mamba frozen-backbone output-stage fine-tuning}
\label{alg:output_stage_finetuning}
\small
\begin{algorithmic}[1]
\Require Training, validation, and test sets for one IXI setting; trained backbone and existing Attention Refine Head weights
\Ensure Best-validation checkpoint and final test PSNR/SSIM
\State Load the backbone and Attention Refine Head weights
\State Fix $\theta_B$; set trainable parameters $\Omega=\{\phi,\psi\}$
\State Configure AdamW with base/minimum learning rates $10^{-3}/10^{-6}$ and 100 warmup iterations
\State $P_{\mathrm{best}}\gets-\infty$
\For{$t=1,\ldots,5000$}
  \State Sample training batch $\mathcal{B}$ of size 4
  \For{each $(\mathbf{x}_{us,j},\mathbf{M}_j,\mathbf{C}_j,\mathbf{x}_{gt,j})$ in $\mathcal{B}$}
    \State $\mathbf{x}_{m,j}\gets B_{\theta_B}(\mathbf{x}_{us,j},\mathbf{M}_j,\mathbf{C}_j)$
    \State $\mathbf{x}_{a,j}\gets A_\phi(\mathbf{x}_{m,j})$
    \State $\hat{\mathbf{x}}_j\gets G_\psi(\mathbf{x}_{a,j})$
    \State $\hat{\mathbf{m}}_j\gets\sqrt{(\hat{\mathbf{x}}_j^{(r)})^2+(\hat{\mathbf{x}}_j^{(i)})^2}$
  \EndFor
  \State $\mathcal{L}\gets\frac{1}{|\mathcal{B}|HW}\sum_{j=1}^{|\mathcal{B}|}
      \|\hat{\mathbf{m}}_j-\mathbf{x}_{gt,j}\|_1$
  \State Update only $\phi,\psi$ using AdamW; keep $\theta_B$ fixed
  \If{$t\bmod500=0$}
    \State Evaluate magnitude-domain validation PSNR $P_{\mathrm{val}}$
    \If{$P_{\mathrm{val}}>P_{\mathrm{best}}$}
      \State Save the best checkpoint; $P_{\mathrm{best}}\gets P_{\mathrm{val}}$
    \EndIf
  \EndIf
  \If{$t\bmod1000=0$}
    \State Save a regular checkpoint
  \EndIf
\EndFor
\State Reload the best-validation checkpoint
\State Evaluate the test set using magnitude reconstructions
\State Report test PSNR and $100\times\mathrm{SSIM}$
\end{algorithmic}
\end{algorithm}

\section{Experimental Setup and Results Analysis}

\subsection{Dataset}

The effectiveness of the proposed AGGR-Mamba is evaluated on the IXI brain MRI dataset. The IXI dataset was collected as part of the IXI--Information eXtraction from Images project~\cite{26}, and is publicly available at: \url{https://brain-development.org/ixi-dataset/}. The dataset contains multiple MRI sequences from healthy subjects, including T1-weighted, T2-weighted, and PD-weighted images. This study focuses on single-modality undersampled brain MRI reconstruction and compares the proposed method with ReconFormer and other representative baselines. Experiments are conducted on both T1-weighted and T2-weighted modalities, with training and testing performed under the $R=4$ and $R=8$ undersampling settings.

All experiments in this work are conducted on a server equipped with a single NVIDIA Tesla V100-PCIE-32GB GPU for training and testing. The model is implemented with PyTorch 2.0.1+cu118 and Python 3.8.20. The experiments use the IXI single-coil brain MRI data that have been preprocessed into H5 format. In the processed H5 data used in this work, the T1 setting contains 2950 training slices, 600 validation slices, and 1200 testing slices, while the T2 setting contains 2443 training slices, 512 validation slices, and 1040 testing slices. The four experimental settings, namely IXI-T1 $R=4$, IXI-T1 $R=8$, IXI-T2 $R=4$, and IXI-T2 $R=8$, are trained and tested independently.

During training, only the training set is used to update the output-stage parameters. The model is evaluated on the validation set every 500 iterations, and the best model weights are saved according to validation PSNR. The final testing metrics are computed on the test set using the corresponding best model weights. The test set is not involved in model selection. The model output is represented by two real/imaginary channels and is converted into a magnitude image before PSNR and SSIM are calculated. The SSIM(\%) reported in the tables is defined as $\mathrm{SSIM}\times100$.

\subsection{Implementation Details}

The model used in the main experiments is AGGR-Mamba. Built on the Mamba backbone, AGGR-Mamba further incorporates an Attention Refine Head and a GDFN-style gated residual refinement branch at the output stage. In the final optimization stage, the trained backbone is frozen and only the Attention Refine Head and GDFN-style branch are updated. These modules refine the backbone reconstruction through local feature recalibration and gated residual compensation.

The comparison methods include UNet~\cite{10}, E2E-VarNet~\cite{21}, SwinUNET~\cite{27}, SwinMR~\cite{28}, and ReconFormer~\cite{22}. Among them, UNet serves as a representative CNN-based reconstruction baseline, E2E-VarNet represents physics-guided unrolled reconstruction methods, SwinUNET and SwinMR are included as Transformer-based reconstruction methods, and ReconFormer represents recurrent Transformer-based MRI reconstruction. By covering CNNs, unrolled networks, Transformers, and recurrent Transformer models, these comparisons provide a more comprehensive evaluation of AGGR-Mamba on undersampled brain MRI reconstruction using the IXI dataset.

\begin{figure*}[!t]
\centering
\includegraphics[
width=0.98\textwidth,
trim=0 0 0 0,
clip
]{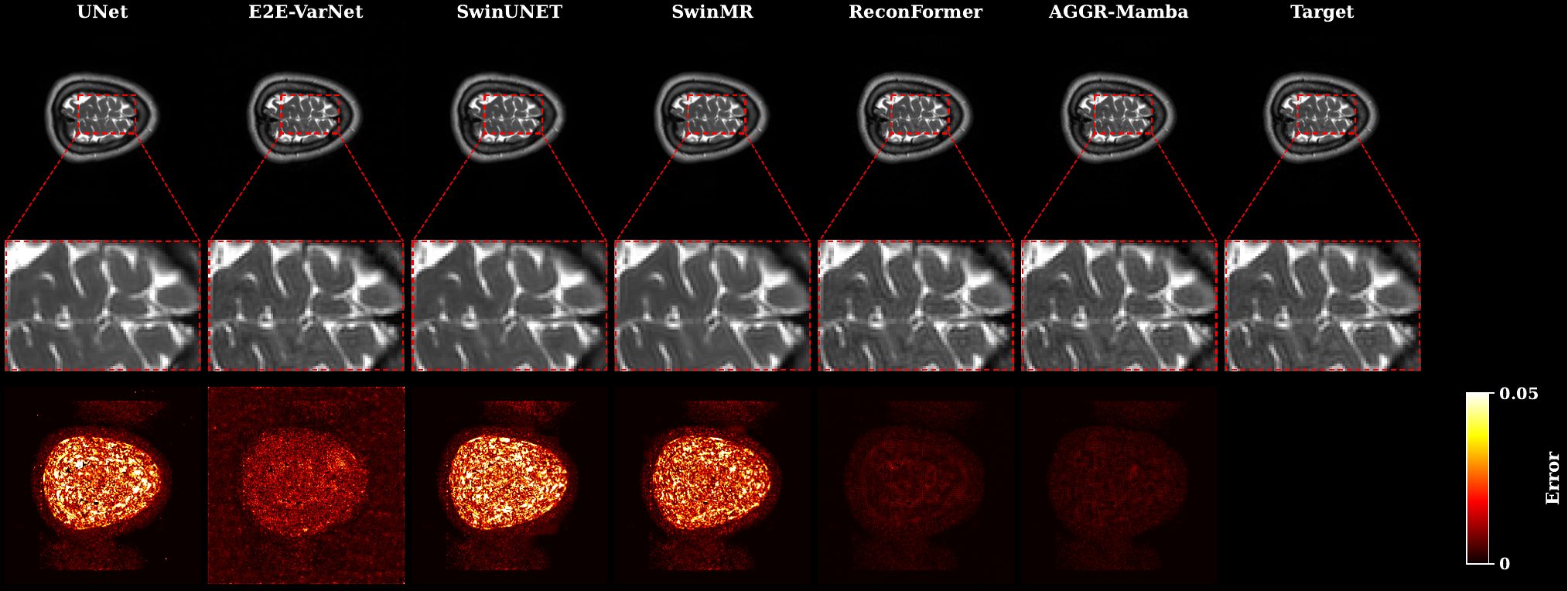}
\caption{Reconstruction results and error maps of different methods on IXI-T2 $R=4$. The zoomed-in regions and error maps are placed below each reconstructed image for visual comparison.}
\label{fig:ixi_t2_r4_visual_comparison}
\end{figure*}

\vspace{0.8em}

\begin{table*}[!t]
\centering
\caption{Quantitative results of different methods on the IXI dataset. The best performance is marked in bold, while the second-best performance is underlined.}
\label{tab:ixi_quantitative_comparison}
\small
\setlength{\tabcolsep}{4pt}
\renewcommand{\arraystretch}{1.15}
\begin{tabular*}{\textwidth}{@{\extracolsep{\fill}}lcccccccc@{}}
\toprule
\multirow{2}{*}{Method}
& \multicolumn{2}{c}{IXI-T1 $R=4$}
& \multicolumn{2}{c}{IXI-T1 $R=8$}
& \multicolumn{2}{c}{IXI-T2 $R=4$}
& \multicolumn{2}{c}{IXI-T2 $R=8$} \\
\cmidrule(lr){2-3}\cmidrule(lr){4-5}\cmidrule(lr){6-7}\cmidrule(lr){8-9}
& PSNR$\uparrow$ & SSIM(\%)$\uparrow$
& PSNR$\uparrow$ & SSIM(\%)$\uparrow$
& PSNR$\uparrow$ & SSIM(\%)$\uparrow$
& PSNR$\uparrow$ & SSIM(\%)$\uparrow$ \\
\midrule
UNet & 37.01 & 94.07 & 31.82 & 90.57 & 38.06 & 94.85 & 33.86 & 91.98 \\
E2E-VarNet & \underline{41.94} & 98.56 & \textbf{35.72} & \underline{97.21} & 42.96 & 98.69 & 37.23 & 97.34 \\
SwinUNET & 34.77 & 94.73 & 30.59 & 91.75 & 37.54 & 95.11 & 33.92 & 92.44 \\
SwinMR & 38.30 & 96.05 & 33.76 & 93.65 & 40.03 & 96.17 & 35.75 & 93.89 \\
ReconFormer & \textbf{42.36} & \textbf{99.24} & 33.87 & \textbf{97.51} & \underline{46.71} & \textbf{99.53} & \underline{38.34} & \textbf{98.45} \\
\textbf{AGGR-Mamba (Ours)} & \underline{41.94} & \underline{99.11} & \underline{35.39} & 97.04 & \textbf{46.99} & \underline{99.45} & \textbf{38.48} & \underline{97.88} \\
\bottomrule
\end{tabular*}
\end{table*}

\subsection{Quantitative Analysis}

Table~\ref{tab:ixi_quantitative_comparison} reports the quantitative comparison of different methods on the IXI-T1 and IXI-T2 modalities under the $R=4$ and $R=8$ undersampling settings. Overall, AGGR-Mamba achieves competitive results across all four testing settings, with particularly stable performance on the T2-weighted reconstruction tasks. For IXI-T2 $R=4$, AGGR-Mamba reaches 46.99 dB in PSNR, higher than the 46.71 dB achieved by ReconFormer, and ranks first among all methods in terms of PSNR. Its SSIM(\%) is 99.45, which is slightly below ReconFormer's 99.53. For IXI-T2 $R=8$, AGGR-Mamba again reports the best PSNR, with 38.48 dB versus 38.34 dB for ReconFormer. This result suggests that, for the T2 modality, output-stage attention recalibration and gated residual compensation can effectively refine the initial reconstruction produced by the Mamba backbone.

AGGR-Mamba also maintains a strong performance on the IXI-T1 tasks. Under the IXI-T1 $R=4$ setting, it obtains a PSNR of 41.94 dB and an SSIM(\%) of 99.11, which are close to ReconFormer's 42.36 dB and 99.24. In the IXI-T1 $R=8$ setting, E2E-VarNet achieves the best PSNR, while ReconFormer gives the best SSIM. AGGR-Mamba obtains a PSNR of 35.39 dB, outperforming ReconFormer at 33.87 dB and ranking second among all compared methods. These results suggest that AGGR-Mamba remains stable across different modalities and undersampling rates. Its advantage is more evident in PSNR improvement and residual error correction, although the structural similarity metric in some T1 settings still leaves room for further improvement.

From the perspective of model types, UNet and SwinUNET generally obtain lower quantitative scores, which suggests that pure CNN reconstruction and directly transferred U-shaped Transformer designs remain limited on the current IXI task. SwinMR improves over UNet and SwinUNET in most cases, but its performance is still behind ReconFormer and AGGR-Mamba. E2E-VarNet performs well on IXI-T1 $R=8$, reflecting the benefit of physics-guided unrolled networks in enforcing data consistency under certain high-acceleration settings. By comparison, AGGR-Mamba first produces a stable reconstruction through the Mamba backbone and then improves PSNR with output-stage attention recalibration and gated residual compensation. Its SSIM remains second-best or close to the best result in several settings, suggesting that the proposed lightweight refinement design provides useful compensation for local reconstruction errors.

\subsection{Analysis of Experimental Results}

Figure~\ref{fig:ixi_t2_r4_visual_comparison} shows the reconstructed images, zoomed-in regions, and error maps of different methods under the IXI-T2 $R=4$ setting. As can be seen, the error maps of UNet, SwinUNET, and SwinMR still contain noticeable bright regions, indicating that residual errors remain around local textures and edges. E2E-VarNet and ReconFormer produce lower errors, and their reconstructed images are closer to the target image in terms of overall structure. By contrast, the error map of AGGR-Mamba is generally darker, and the brain tissue boundaries and texture patterns in the zoomed-in regions are also clearer. Its advantage is mainly reflected in reduced error-map intensity and more stable local structures. This suggests that the output-stage refinement modules further correct the initial reconstruction produced by the Mamba backbone, rather than improving the metrics by substantially changing the global image contrast.

\subsection{Ablation Study Analysis}

To assess the effect of the output-stage refinement modules, we perform ablation experiments on IXI-T2 $R=4$ and further compare the late-stage GDFN fine-tuning strategy with joint fine-tuning. All reported results are evaluated on the same fixed test set, with the best checkpoint selected based on validation PSNR.

\begin{table}[t]
\centering
\caption{Module-wise ablation study on IXI-T2 $R=4$. $\Delta$ indicates the PSNR gain relative to the M\textnormal{amba} backbone.}
\label{tab:module_ablation_t2r4}
\scriptsize
\setlength{\tabcolsep}{3pt}
\renewcommand{\arraystretch}{1.12}
\resizebox{\columnwidth}{!}{
\begin{tabular}{lcccccc}
\toprule
Variant & Attention Refine & GDFN & Joint Fine-tuning & PSNR(dB)$\uparrow$ & SSIM(\%)$\uparrow$ & $\Delta$PSNR \\
\midrule
Mamba Backbone & $\times$ & $\times$ & $\times$ & 46.4907 & 99.38 & -- \\
Attention Refine & $\checkmark$ & $\times$ & $\times$ & 46.9393 & 99.44 & +0.4486 \\
Attention Refine + GDFN & $\checkmark$ & $\checkmark$ & $\times$ & 46.9930 & 99.44 & +0.5023 \\
AGGR-Mamba & $\checkmark$ & $\checkmark$ & $\checkmark$ & 46.9945 & 99.45 & +0.5038 \\
\bottomrule
\end{tabular}}
\end{table}

\begin{figure}[!t]
\centering
\includegraphics[width=\columnwidth]{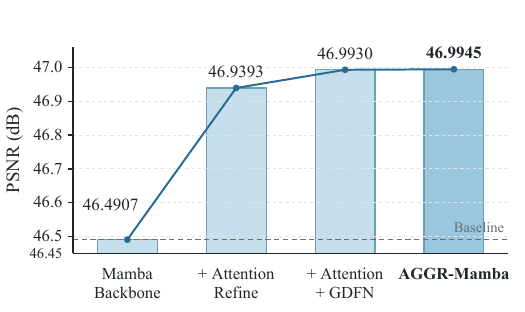}
\caption{PSNR comparison of different module combinations on IXI-T2 $R=4$.}
\label{fig:psnr_improvement_t2r4}
\end{figure}

Table~\ref{tab:module_ablation_t2r4} reports the module-wise ablation results under the IXI-T2 $R=4$ setting. The baseline Mamba backbone obtains a PSNR of 46.4907 dB and an SSIM of 99.38\%. Adding the Attention Refine Head raises the PSNR to 46.9393 dB, which shows that output-stage attention refinement helps recover local details in the initial reconstruction. When the GDFN-style gated residual branch is further introduced, the PSNR improves to 46.9930 dB, suggesting that the gated residual selection mechanism provides additional compensation for the remaining reconstruction errors. With output-stage joint fine-tuning, the final AGGR-Mamba reaches 46.9945 dB in PSNR and 99.45\% in SSIM. The improvement over late-stage GDFN fine-tuning is modest, but the results are more stable overall, indicating that jointly optimizing the Attention Refine Head and the GDFN-style branch benefits the final refinement stage.

\begin{table}[t]
\centering
\caption{Comparison between late-stage GDFN fine-tuning and joint fine-tuning on the four IXI test settings. Each cell reports PSNR(\textnormal{d}B) / SSIM(\%).}
\label{tab:headft_ablation}
\resizebox{\columnwidth}{!}{
\begin{tabular}{lccc}
\toprule
Setting & Late-stage GDFN Fine-tuning & Joint Fine-tuning & $\Delta$PSNR \\
\midrule
IXI-T1 $R=4$ & 41.7266 / 99.11 & 41.9360 / 99.11 & +0.2094 \\
IXI-T1 $R=8$ & 35.3777 / 97.04 & 35.3879 / 97.04 & +0.0102 \\
IXI-T2 $R=4$ & 46.9930 / 99.44 & 46.9945 / 99.45 & +0.0015 \\
IXI-T2 $R=8$ & 38.4408 / 97.87 & 38.4789 / 97.88 & +0.0381 \\
\bottomrule
\end{tabular}}
\end{table}

Table~\ref{tab:headft_ablation} compares the two output-stage training strategies. Joint fine-tuning yields higher PSNR than late-stage GDFN fine-tuning in all four evaluated settings, with gains ranging from 0.0015 to 0.2094 dB. These differences describe the reported checkpoints and do not establish statistical significance.

\section{Discussion and Conclusion}

AGGR-Mamba refines the output of a trained Mamba reconstruction backbone through channel-spatial attention and GDFN-style gated residual filtering. Final-stage optimization updates only the two refinement modules while keeping the backbone fixed. On the evaluated IXI-T2 settings, the method achieves PSNR values of 46.99 dB at $R=4$ and 38.48 dB at $R=8$. The IXI-T2 $R=4$ ablation identifies Attention Refine as the main source of improvement, with smaller additional gains from GDFN and joint fine-tuning.

The findings are limited to the evaluated IXI single-coil brain MRI settings. Performance is more favorable on T2 images, and the method does not achieve the highest SSIM in every setting. The small differences between fine-tuning strategies require further validation before drawing conclusions about stability. Future work will investigate realistic multi-coil reconstruction and refinement designs incorporating frequency-domain constraints.


\begin{thebibliography}{28}

\bibitem{1}
K. P. Pruessmann, M. Weiger, M. B. Scheidegger, and P. Boesiger, ``SENSE: Sensitivity encoding for fast MRI,'' \emph{Magnetic Resonance in Medicine}, vol. 42, no. 5, pp. 952--962, 1999, doi: \url{https://doi.org/10.1002/(SICI)1522-2594(199911)42:5<952::AID-MRM16>3.0.CO;2-S}.

\bibitem{2}
M. A. Griswold, P. M. Jakob, R. M. Heidemann, M. Nittka, V. Jellus, J. Wang, B. Kiefer, and A. Haase, ``Generalized autocalibrating partially parallel acquisitions (GRAPPA),'' \emph{Magnetic Resonance in Medicine}, vol. 47, no. 6, pp. 1202--1210, 2002, doi: 10.1002/mrm.10171.

\bibitem{3}
M. Lustig, D. Donoho, and J. M. Pauly, ``Sparse MRI: The application of compressed sensing for rapid MR imaging,'' \emph{Magnetic Resonance in Medicine}, vol. 58, no. 6, pp. 1182--1195, 2007, doi: 10.1002/mrm.21391.

\bibitem{4}
E. J. Cand\`{e}s, J. Romberg, and T. Tao, ``Robust uncertainty principles: Exact signal reconstruction from highly incomplete frequency information,'' \emph{IEEE Transactions on Information Theory}, vol. 52, no. 2, pp. 489--509, 2006, doi: 10.1109/TIT.2005.862083.

\bibitem{5}
D. L. Donoho, ``Compressed sensing,'' \emph{IEEE Transactions on Information Theory}, vol. 52, no. 4, pp. 1289--1306, 2006, doi: 10.1109/TIT.2006.871582.

\bibitem{6}
M. Lustig, D. L. Donoho, J. M. Santos, and J. M. Pauly, ``Compressed sensing MRI,'' \emph{IEEE Signal Processing Magazine}, vol. 25, no. 2, pp. 72--82, Mar. 2008, doi: 10.1109/MSP.2007.914728.

\bibitem{7}
D. Liang, J. Cheng, Z. Ke, and L. Ying, ``Deep MRI reconstruction: Unrolled optimization algorithms meet neural networks,'' arXiv preprint arXiv:1907.11711, 2020.

\bibitem{8}
S. Wang, Z. Su, L. Ying, X. Peng, S. Zhu, F. Liang, D. Feng, and D. Liang, ``Accelerating magnetic resonance imaging via deep learning,'' in \emph{Proc. IEEE International Symposium on Biomedical Imaging (ISBI)}, 2016, pp. 514--517, doi: 10.1109/ISBI.2016.7493320.

\bibitem{9}
B. Zhu, J. Z. Liu, B. R. Rosen, and M. S. Rosen, ``Image reconstruction by domain-transform manifold learning,'' \emph{Nature}, vol. 555, pp. 487--492, 2018, doi: 10.1038/nature25988.

\bibitem{10}
O. Ronneberger, P. Fischer, and T. Brox, ``U-Net: Convolutional networks for biomedical image segmentation,'' in \emph{Medical Image Computing and Computer-Assisted Intervention--MICCAI 2015}, LNCS, vol. 9351, pp. 234--241, 2015, doi: \url{https://doi.org/10.1007/978-3-319-24574-4_28}.

\bibitem{11}
I. Goodfellow \emph{et al.}, ``Generative adversarial nets,'' in \emph{Advances in Neural Information Processing Systems (NeurIPS)}, 2014, pp. 2672--2680.

\bibitem{12}
G. Yang \emph{et al.}, ``DAGAN: Deep de-aliasing generative adversarial networks for fast compressed sensing MRI reconstruction,'' \emph{IEEE Transactions on Medical Imaging}, vol. 37, no. 6, pp. 1310--1321, 2018, doi: 10.1109/TMI.2017.2785879.

\bibitem{13}
A. Vaswani \emph{et al.}, ``Attention is all you need,'' in \emph{Advances in Neural Information Processing Systems (NeurIPS)}, 2017, pp. 5998--6008.

\bibitem{14}
Z. Liu \emph{et al.}, ``Swin Transformer: Hierarchical vision transformer using shifted windows,'' in \emph{Proc. IEEE/CVF International Conference on Computer Vision (ICCV)}, 2021, pp. 10012--10022, doi: 10.1109/ICCV48922.2021.00986.

\bibitem{15}
A. Gu and T. Dao, ``Mamba: Linear-time sequence modeling with selective state spaces,'' in \emph{International Conference on Learning Representations (ICLR)}, 2024, arXiv:2312.00752, doi: 10.48550/arXiv.2312.00752.

\bibitem{16}
Y. Liu \emph{et al.}, ``VMamba: Visual state space model,'' in \emph{Advances in Neural Information Processing Systems (NeurIPS)}, 2024, arXiv:2401.10166.

\bibitem{17}
Y. Korkmaz and V. M. Patel, ``MambaRecon: MRI reconstruction with structured state space models,'' in \emph{Proc. IEEE/CVF Winter Conference on Applications of Computer Vision (WACV)}, 2025, doi: 10.48550/arXiv.2409.12401.

\bibitem{18}
K. He, X. Zhang, S. Ren, and J. Sun, ``Deep residual learning for image recognition,'' in \emph{Proc. IEEE Conference on Computer Vision and Pattern Recognition (CVPR)}, 2016, pp. 770--778, doi: 10.1109/CVPR.2016.90.

\bibitem{19}
J. Schlemper, J. Caballero, J. V. Hajnal, A. Price, and D. Rueckert, ``A deep cascade of convolutional neural networks for dynamic MR image reconstruction,'' \emph{IEEE Transactions on Medical Imaging}, vol. 37, no. 2, pp. 491--503, 2018, doi: 10.1109/TMI.2017.2760978.

\bibitem{20}
K. Hammernik \emph{et al.}, ``Learning a variational network for reconstruction of accelerated MRI data,'' \emph{Magnetic Resonance in Medicine}, vol. 79, no. 6, pp. 3055--3071, 2018, doi: 10.1002/mrm.26977.

\bibitem{21}
A. Sriram \emph{et al.}, ``End-to-end variational networks for accelerated MRI reconstruction,'' in \emph{Medical Image Computing and Computer-Assisted Intervention--MICCAI 2020}, pp. 64--73, 2020, doi: \url{https://doi.org/10.1007/978-3-030-59713-9_7}.

\bibitem{22}
P. Guo, Y. Mei, J. Zhou, S. Jiang, and V. M. Patel, ``ReconFormer: Accelerated MRI reconstruction using recurrent Transformer,'' \emph{IEEE Transactions on Medical Imaging}, vol. 43, no. 1, pp. 582--593, 2024, doi: 10.1109/TMI.2023.3314747.

\bibitem{23}
S. W. Zamir, A. Arora, S. Khan, M. Hayat, F. S. Khan, and M.-H. Yang, ``Restormer: Efficient Transformer for high-resolution image restoration,'' in \emph{Proc. IEEE/CVF Conference on Computer Vision and Pattern Recognition (CVPR)}, 2022, pp. 5718--5729, doi: 10.1109/CVPR52688.2022.00564.

\bibitem{24}
H. R. Sheikh, M. F. Sabir, and A. C. Bovik, ``A statistical evaluation of recent full reference image quality assessment algorithms,'' \emph{IEEE Transactions on Image Processing}, vol. 15, no. 11, pp. 3440--3451, Nov. 2006, doi: 10.1109/TIP.2006.881959.

\bibitem{25}
Z. Wang, A. C. Bovik, H. R. Sheikh, and E. P. Simoncelli, ``Image quality assessment: From error visibility to structural similarity,'' \emph{IEEE Transactions on Image Processing}, vol. 13, no. 4, pp. 600--612, Apr. 2004, doi: 10.1109/TIP.2003.819861.

\bibitem{26}
IXI Dataset, ``Information eXtraction from Images,'' available online: \url{https://brain-development.org/ixi-dataset/}.

\bibitem{27}
H. Cao \emph{et al.}, ``Swin-Unet: Unet-like pure Transformer for medical image segmentation,'' arXiv preprint arXiv:2105.05537, 2021, doi: 10.48550/arXiv.2105.05537.

\bibitem{28}
J. Huang, Y. Fang, Y. Wu, H. Wu, Z. Gao, Y. Li, J. Del Ser, J. Xia, and G. Yang, ``Swin Transformer for fast MRI,'' \emph{Neurocomputing}, vol. 493, pp. 281--304, 2022, doi: 10.1016/j.neucom.2022.04.051.

\end{thebibliography}
\end{document}